\documentclass{article}

\usepackage{arxiv}

\usepackage[utf8]{inputenc} 
\usepackage[T1]{fontenc}    
\usepackage{hyperref}       
\usepackage{url}            
\usepackage{booktabs}       
\usepackage{amsfonts}       
\usepackage{nicefrac}       
\usepackage{microtype}      
\usepackage{graphicx}
\usepackage{natbib}
\usepackage{doi}

\graphicspath{ {./figures/} }
\usepackage{verbatim} 

\usepackage{paralist}
\usepackage{booktabs}
\usepackage{caption}
\usepackage{subcaption}
\usepackage{amsmath}
\usepackage{cleveref}
\usepackage[none]{hyphenat}
\usepackage{mathtools}
\DeclarePairedDelimiter{\ceil}{\lceil}{\rceil}

\title{Spoiler in a Textstack: How Much Can Transformers Help?}

\author{ Anna Wróblewska\\
	Faculty of Mathematics and Information Science \\
	Warsaw University of Technology\\
	Warsaw, Poland\\
	\texttt{anna.wroblewska1@pw.edu.pl}
	\And
	Paweł Rzepiński\\
	Faculty of Mathematics and Information Science \\
	Warsaw University of Technology\\
	Warsaw, Poland\\
	\texttt{rzepinski.pawel@gmail.com}\\
	\And
	Sylwia Sysko-Romańczuk \\
	Faculty of Management \\
	Warsaw University of Technology\\
	Warsaw, Poland \\
	\texttt{sylwia.sysko.romanczuk@pw.edu.pl}\\
}

\date{}

\renewcommand{\shorttitle}{Spoiler in a Textstack}

\hypersetup{
pdftitle={Spoiler in a Textstack: How Much Can Transformers Help?},
pdfsubject={CS.CL, cs.LG},
pdfauthor={Anna Wróblewska, Paweł Rzepiński, Sylwia Sysko-Romańczuk},
pdfkeywords={Spoiler Annotations, Deep Learning,  Natural Language Processing, Interpretability},
}

\begin{document}
\maketitle

\begin{abstract}
This paper presents our research regarding spoiler detection in reviews. In this use case, we describe the method of fine-tuning and organizing the available methods of text-based model tasks with the latest deep learning achievements and techniques to interpret the models' results.

Until now, spoiler research has been rarely described in the literature. We tested the transfer learning approach and different latest transformer architectures on two open datasets with annotated spoilers (ROC AUC above 81\% on TV Tropes Movies dataset, and Goodreads dataset above 88\%). We also collected data and assembled a new dataset with fine-grained annotations. To that end, we employed interpretability techniques and measures to assess the models' reliability and explain their results. 	
\end{abstract}


\section{Introduction}
\label{introduction}

Reviews and comments about a movie play a significant role in its popularity. Before deciding whether to watch a movie or show, potential viewers read many reviews. Accidentally reading information that uncovers essential parts of the plot -- called "spoilers" -- can ruin the excitement, surprise, and reduce the number of people who wish to watch the movie. Finally, the consequences of the drop in viewership can significantly impact the return on investment of affiliate producers and distributors. The degree of plot uncertainty resolved by movie reviews (i.e., spoiler intensity) has a positive and significant association with box office revenue~\citep{Ryoo_2020}. The effective detection of spoilers is an important technological and business task.

\begin{figure}[htb]
    \centering
    \frame{\includegraphics[width=0.6\linewidth]{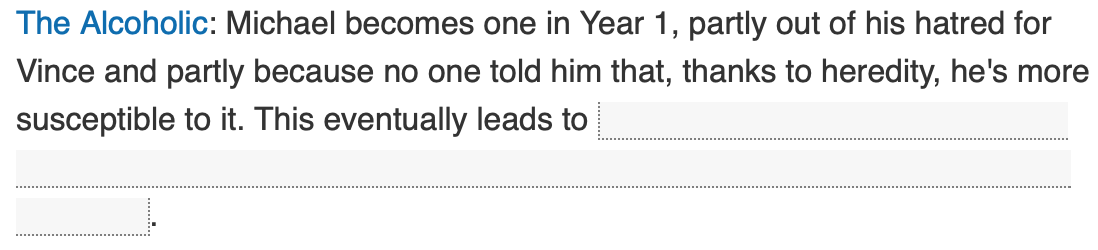}}
    \caption{Example of an entry hiding a tagged spoiler. Source page available at \small{\url{https://tvtropes.org/pmwiki/pmwiki.php/Literature/SuperPowereds}}.}
    \label{fig:tvtropes-spoiler_example}
\end{figure}

A spoiler is a piece of information regarding a media product (i.e., a book, a movie) that reveals crucial details of the plot.
The automatic spoiler detection system is used to lower review portals moderation effort to keep the information "spoiler-free." It automatically flags fragments of texts as spoilers and redirects suspicious content to human moderators for review. The system can also remind the author about tagging detected spoilers before posting the content. 
Due to the context- and time-dependent nature of spoilers, their automatic detection is not an easy task (see the example in~Figure~\ref{fig:tvtropes-spoiler_example}). 

{}

Currently, novel deep learning architectures -- Transformers~\citep{wolf-etal-2020-transformers-m} -- achieve high performance in language modeling and NLP tasks. They work on an acceptable level even for context-sensitive data that requires some reasoning on texts having general knowledge about the world. These features also characterize spoiler reviews.

Spoiler detection usually works on a document or sentence level. The former involves tagging the entire text as a spoiler, while the latter restricts the tagging to one sentence or even particular spoiler phrases. The sentence-level solution is more desirable than the document one because it can be easily extended to documents, removing much of the ambiguity about spoilers' location. 

Designing our study, we addressed the following research questions. 1. Can the latest deep learning architectures~\citep{cohan_pretrained_2019,sun_how_2020,wolf-etal-2020-transformers-m} improve performance metrics of sentence-level spoiler detection? 2. How can we check and measure whether the results are reliable? We also proposed the architecture extensions to deal with additional genre information. We investigated whether and how the deep learning models understood the data: on which parts of texts they concentrate to make their model decisions. We measured this with an approach for measuring model understanding and interpretability techniques~\citep{deyoung_eraser_2020}.

In studies on spoilers up to date, the task of spoiler detection was conducted using algorithms trained from scratch~\citep{golbeck_twitter_2012,iwai_sentence-based_2014-1,ueno_spoiler_2019,chang_deep_2018,wan_fine-grained_2019}. The novelty of our approach is to use the transfer learning approach to utilize the knowledge the algorithm gained during training for one task when solving another that is somehow related. 

Transformer architectures achieved state-of-the-art results in many NLP tasks~\citep{devlin_bert_2019,brunner_identifiability_2020,wolf-etal-2020-transformers-m}. They proved their effectiveness with the transfer learning approach, moving from one task to another.
Moreover, the underlying Transformers can perform multiple computations in parallel, hence achieving huge speed-ups from the specialized hardware (TPUs)~\citep{vaswani_attention_2017}. 

Many papers report results on custom datasets that are not published or too specific to reuse for spoiler detection. Currently, there are only two datasets for spoiler detection tasks with sentence-level labels and more than 10,000 entries~\citep{wan_fine-grained_2019,boydgraber_spoiler_2013}.

In this article, we present the following research approach:
\begin{enumerate}
    \item We examine the transfer-learning approach to spoiler detection task: \\ fine-tuning language-modelling neural networks with Transformer (see Section~\ref{sec:method} and experiments in Section~\ref{sec:experiments}). 
    \item We propose two different ways to incorporate book genres to the model which leads to results improvement (see Subsection~\ref{sec:method-append} and experiments in Section~\ref{sec:experiments}).
    \item We check the reliability of models with novel interpretability techniques (see Section~\ref{sec:analysis}).
\end{enumerate}


Our contribution is also publishing a new dataset \textit{TV Tropes Books} with a unique characteristic: word-level spoiler annotations (see Subsection~\ref{sec:our-dataset}). Thanks to the new dataset, the number of spoiler-detection applications will increase the number of studies on the relationship between spoilers and demand, filling the literature research gap.


\section{Related Work} \label{ch:spoiler_task_history}

Initially, spoiler detection was described as a text classification task of assigning binary labels -- contains spoilers or not -- to a given text data (usually a document or a sentence). Earlier studies recommended filtering the spoilers by simple keyword matching~\citep{golbeck_twitter_2012, nakamura_temporal_2007}.

\citet{hijikata_context-based_2016} and~\citet{iwai_sentence-based_2014} also used the bag-of-words approach to represent text as features later used by the Support Vector Machine (SVM) or the Naive Bayes algorithm in order to classify review comments on the Amazon platform. In~\citet{hijikata_context-based_2016}, additional information like sentence location and spoiler probabilities for the adjacent sentences were leveraged to increase accuracy (F1 scores).

The method presented in~\citet{boydgraber_spoiler_2013} incorporated non-text features into the model, e.g. a movie's genre, length, and airing date. Hopper and Jeon with colleagues highlighted two linguistic aspects related to spoilers: transitivity and narrativeness~\citep{hopper_transitivity_1980,jeon_spoiler_2016}. They noticed that, for example, more named entities, objectivity, special verbs, and tenses are used in narrative texts. 

Current state-of-the-art approaches use neural networks with word embeddings as input, hence mostly removing the need for hand-crafted text features. \citet{chang_deep_2018} designed a model leveraging GloVe\footnote{\url{https://nlp.stanford.edu/projects/glove/}}
for word representations, Gated Recurrent Units (GRU) for the textual input, and Convolutional Neural Network (CNN) for encoding the movie genre information. That solution achieved the accuracy of 75\% (an increase from 64\%) on the dataset published alongside \citep{boydgraber_spoiler_2013}. Similar techniques were employed for Japanese-written reviews by~\citet{ueno_spoiler_2019} who improved the previous solution that did not use neural networks~\citep{hijikata_context-based_2016}. \citet{ueno_spoiler_2019} used the Long-Short Term Memory (LSTM) model to leverage sequential characteristics of the data and utilized fastText library\footnote{\url{https://github.com/facebookresearch/fastText}}
 library to represent the words in the preprocessed text. Cumulatively, these two changes lead to a significant increase in F1 metric (55\% vs. 39\%). \citet{wan_fine-grained_2019} study in spoiler detection also based on word embeddings as a form of an input representation, but it implemented a modified version of the hierarchical attention network described in~\citet{yang_hierarchical_2016}.


\section{Datasets} \label{ch:datasets}

In our experiments we utilized following datasets: Goodreads (with our additional balanced version)~\citep{wan_fine-grained_2019}, TV Tropes Movies~\citep{boydgraber_spoiler_2013}, and prepared by ourselves fine-grained dataset TV Tropes Books. Table~\ref{tab:datasets-summary} summarizes their detailed information.

\begin{table*}[!htb]
\centering
\begin{tabular}{p{2.7cm}ccccc}
\toprule
Dataset            & Docs & Sent. & \begin{tabular}[c]{c}Spoiler\\ sent.\end{tabular} & \begin{tabular}[c]{c}Spoiler \\ sent. {[}\%{]}\end{tabular} & \begin{tabular}[c]{c}Sent. \\ per doc.\end{tabular} \\ \midrule
goodreads & 1,300k     & 17,000k    & 570k              & 3                          & 13                     \\
goodreads-balanced & 180k      & 3150k     & 570k              & 18                         & 18                     \\
tvtropes-movies & 16k       & 16k       & 8k                & 50                         & 1                      \\
tvtropes-books     & 340k      & 670k      & 110k              & 17                         & 2                      \\ \bottomrule
\end{tabular}
\caption{The basic characteristics of the spoiler datasets: values approximated for readability. Note: Docs means a number of documents, Sent. -- a number of sentences, Spoiler sent. -- A number of spoiler sentences, Sent. per doc. -- an average number of sentences per document }
\label{tab:datasets-summary}
\end{table*}

\subsection{Available Spoiler Datasets}

\begin{figure*}[!htb]
\centering
\includegraphics[width=0.45\linewidth]{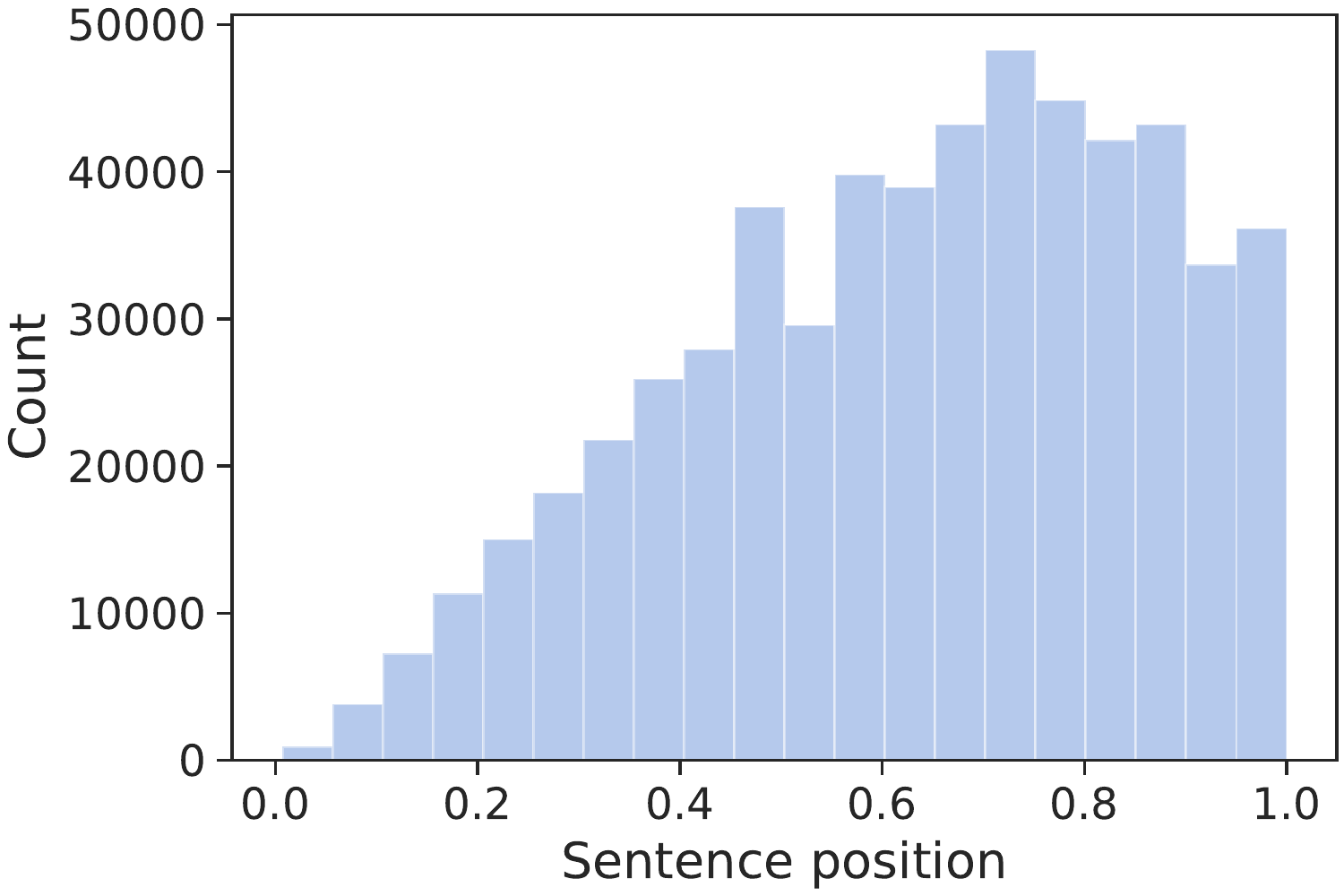}
\caption{The relative normalized position of spoiler sentences in the reviews. Value of $0.0$ represents first position in the review, $1.0$ the last one.
\label{fig:goodreads-spoiler_sentence_position}}
\end{figure*}

\begin{figure*}[!htb]
\centering
\includegraphics[width=0.45\linewidth]{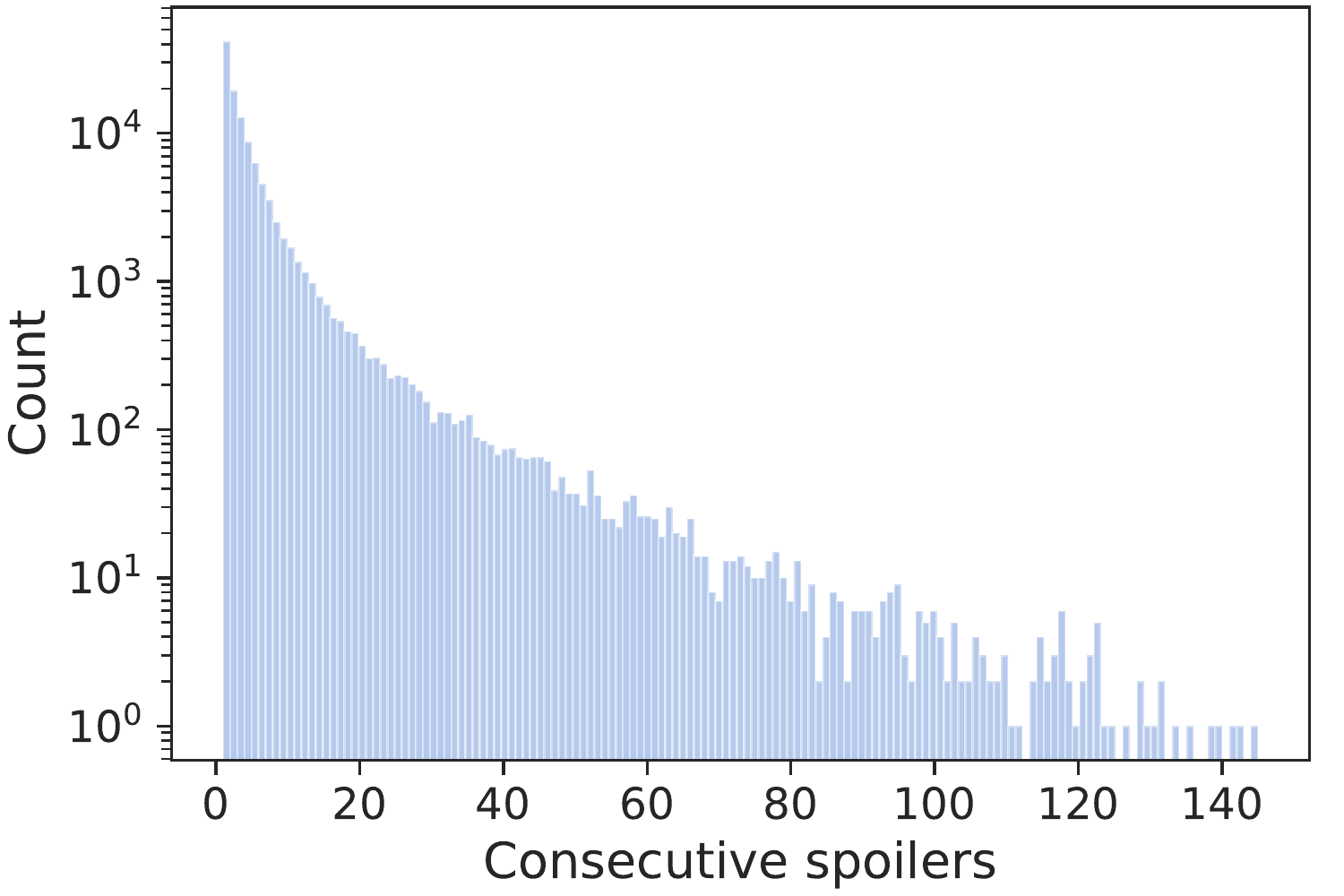}
\caption{The lengths of consecutive spoiler sentences sequences. The Y-axis is in logarithmic scale due to long tail distribution.
\label{fig:goodreads-spoiler_neighbors}}
\end{figure*}

In the Goodreads dataset~\citep{wan_fine-grained_2019},\footnote{\url{https://sites.google.com/eng.ucsd.edu/ucsdbookgraph/reviews}}
 sentences featuring spoilers: (1) tend to appear later in the review text as the distribution of their relative positions in~Figure~\ref{fig:goodreads-spoiler_sentence_position} 
 is not uniform, and (2) often appear together in consecutive sequences (Figure~\ref{fig:goodreads-spoiler_neighbors}); individual spoiler sentences (42,000) constitute only about 7\% of all.
 
\begin{figure*}[!htb]
\centering
\includegraphics[width=0.45\linewidth]{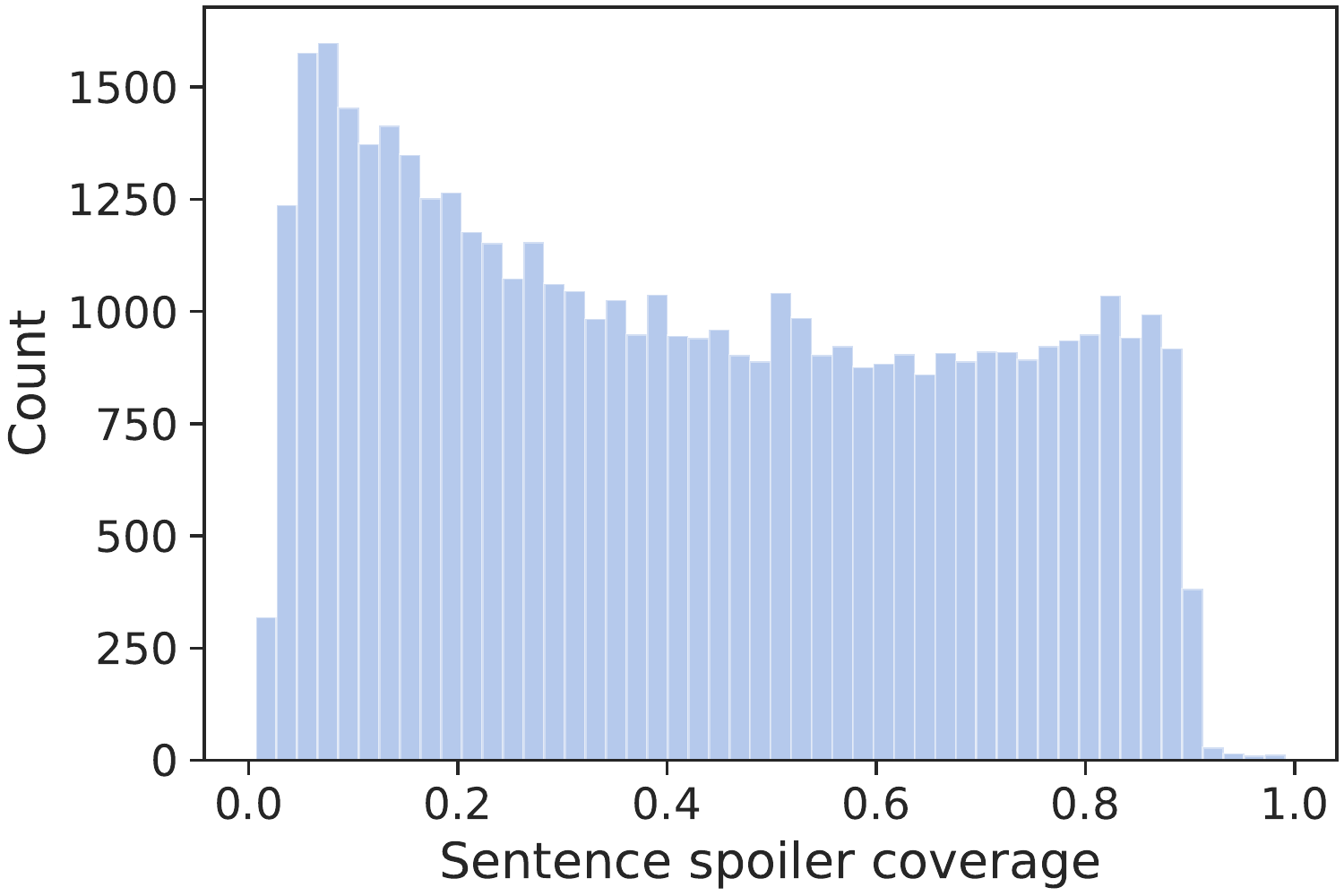}
\caption{The distribution of spoiler sentence portion surrounded with spoiler tags in TV Tropes Books dataset.
\label{fig:spoiler_tags_coverage}}
\end{figure*}

Users may assign custom tags to books on the Goodreads platform and they are often used as a grouping method, e.g. a book genre indicator. The number of such assignments can constitute a score for the particular genre. 
The authors of the dataset selected the subset of the tags related to book genres and merged them into ten distinct groups. \cref{tab:genre-spoilers} presents the data about books grouped by the genres category with the highest score. 
The percentage of the reviews containing spoilers in each genre in the Goodreads dataset ranges from 2.2\% (for the \textit{non-fiction} genre) to 7.5\% (for the \textit{mystery, thriller}, and \textit{crime} genres).

\begin{table}[htb]
\centering
\caption{Percentage of the reviews containing spoilers in each genre group.}
\label{tab:genre-spoilers}
\begin{tabular}{lrr}
\toprule
Genres group                           & 
\begin{tabular}[c]{r}Reviews \\ count \end{tabular} & 
\begin{tabular}[c]{r}Reviews \\ with spoilers [\%] \end{tabular}
 \\ \midrule
non-fiction                            &      23,856 &                2.23 \\
poetry                                 &       1,622 &                2.28 \\
children                               &      15,159 &                4.08 \\
comics, graphic                        &      43,115 &                4.62 \\
romance                                &     264,628 &                5.89 \\
fiction                                &     234,153 &                6.12 \\
history, historical fiction, biography &      21,428 &                6.30 \\
fantasy, paranormal                    &     459,913 &                7.03 \\
young-adult                            &     240,347 &                7.20 \\
mystery, thriller, crime               &      73,811 &                7.49 \\ \bottomrule
\end{tabular}
\end{table}

Due to the size of the Goodreads dataset, we created a document-balanced version (see Table~\ref{tab:datasets-summary}). Our version contains all reviews with at least one spoiler sentence 
and the same number of randomly selected reviews without any spoilers. Because of the different number of sentences in documents, the resulting dataset is not balanced in terms of sentences.

In the TV Tropes Movie dataset~\citep{boydgraber_spoiler_2013}, spoiler boundaries are not clearly defined; these can be a single word, whole sentence, or paragraph. Spoiler annotations are of high quality, because they are reviewed and edited by the whole TV Tropes' community. 

\subsection{New Fine-Grained Dataset}
\label{sec:our-dataset}

Inspired by the quality of spoiler annotations done by the TV Tropes' community -- users defined the boundaries of spoiler fragments -- we prepared a new spoilers' dataset for books that had word-level spoiler annotations. The procedure comprised the four following steps: 
\begin{inparaenum}[(i)]
  \item listing all pages related to books,
  \item archiving HTML contents of the pages,
  \item extracting entries from tropes list,
  \item splitting them into sentences and marking them with spoiler or non-spoiler tags.
\end{inparaenum}
Two first stages of processing came from the MADE recommender project\footnote{\url{https://github.com/raiben/made_recommender}} and follow a similar logic. Then, we parsed the HTML page contents to extract a tropes list. Splitting text into individual sentences is performed using the \textit{spacy} library\footnote{\url{https://spacy.io/}} that leverages a machine learning model built for the English language. 

To leverage word-level spoiler annotations, the output file included:
\begin{inparaenum}[(i)]
  \item the URL of the source page,
  \item the trope associated with the entry,
  \item a boolean indicator whether any sentence has a spoiler,
  \item a triplet consisting of sentence text, sentence label, and a list of spoilers indices. 
\end{inparaenum}

The spoilers for each specific sentence were saved as \texttt{(start, end)} indices pair (with inclusive start and exclusive end) to indicate their positions. As much as 42\% (47,000) of spoiler sentences were annotated at the word level, i.e. with less than 90\% of all characters marked as a spoiler. The average portion of spoiler sentences inside spoiler tags were about 43\% of sentence length. Figure~\ref{fig:spoiler_tags_coverage} depicts the  distribution of spoiler sentences parts surronded with spoiler tags in TV Tropes Books dataset.

The gathered data contained some errors because a spoiler tag could span multiple words or even sentences, and splitting text into sentences was a non-trivial task. We identified more than 30 edge cases and checked them using automatic tests so as to ensure the best possible quality of the dataset.\footnote{Our scripts to scrap the dataset TV Tropes Books: \url{https://github.com/rzepinskip/tvtropes-books}} 

\section{Our Spoiler Detection Models} 
\label{sec:method}

We developed models derived from the fine-tuning method for BERT architecture \citep{devlin_bert_2019} in a few variants differing in: \begin{inparaenum}
  \item the type of the input (single-sentence vs. sequential-sentence classification),
  \item without or with additional encoding (vector of genre probabilities, vector of genre names)
\end{inparaenum}.

\subsection{Single-Sentence Classification (SSC)}

In the SSC variant we examined two variants of classification heads -- networks applied on top of the pre-trained model: 
\begin{enumerate}
  \item the hidden state of last encoder block for the \texttt{CLS} token was used for prediction (\textit{sequence} variant)~\citep{devlin_bert_2019}.
  \item prediction was based on the \texttt{CLS} token embedding from the linear layer during pretraining (\textit{pooled} variant). Here: official BERT code repository  \footnote{\url{https://github.com/google-research/bert}}
   and Transformers library~\citep{wolf-etal-2020-transformers-m}\footnote{\url{https://github.com/huggingface/transformers}}
\end{enumerate}
The final linear layer, preceded by dropout layer, produced a single value passed to the sigmoid activation function. The output was a probability of a positive label for specified input. The initial setting of the model followed the values proposed in~\citet{devlin_bert_2019}. We trained the model for four epochs using the AdamW optimizer with default arguments -- the weight decay rate of $0.01$, $\beta_1 = 0.9$, $\beta_2=0.999$, and $\epsilon=1\text{e-}6$  -- and set the dropout parameter to $0.1$.

The fine-tuned models often suffered from catastrophic forgetting problem, losing the benefits of knowledge gained through language modelling pre-training. To tackle this issue, we employed the slanted triangular learning rate schedule following \citet{howard_universal_2018}. The learning rate increased linearly for 10\% of all training steps (the warm-up phase) and then decreased linearly to 0 (the decay phase).

The spoiler detection problem is a binary classification task. Thus, we utilised the weighted binary cross-entropy as a loss function. To cope with imbalanced sets, we added the additional positive class weighting as dependant on the dataset imbalance ratio with the value of $\frac{\text{negative samples}}{\text{positive samples}}$. Additionally, we tested the focal loss function to research its ability to weight hard-to-predict samples. Faster convergence for imbalanced datasets was achieved by initializing the bias with a constant value of the $log$ of $\frac{\text{positive samples}}{\text{negative samples}}$. 

The hyperparameters were kept unchanged in all experiments. The early stopping with two-epochs prevented unnecessary computation.

\subsection{Sequential-Sentence Classification (SeqSC)} \label{sec:ssc-method}

Spoiler sentences often appear together (as illustrated in~Figure~\ref{fig:goodreads-spoiler_neighbors}). Hence, we utilized a sentence context. We tested the idea of modelling sentence context, inputting into the Transformer model multiple sentences at once.

The number of concatenated sentences depends on technical constraints of the underlying model: for BERT, the maximum input sequence is 512. \citet{cohan_pretrained_2019} propose an algorithm for splitting documents into sentence sequences so as to fit them into the limit. They recursively split the document's sentences into two parts until each split has fewer sentences than the specified threshold. This method can result in uneven splits, e.g. a document with nine sentences and threshold of four results in a \texttt{4,2,3} split. The longest part may be later truncated because of the character limit. This situation can be avoided with more even splits.

Thus, we proposed splitting text evenly, using the \texttt{array\_split} method from the \texttt{numpy} package,
even if this results in a higher number of splits. The number of groups is determined as $\ceil{\frac{\text{sentences count}}{\text{threshold}}}$. The exact value of max sentences per group depends on the average sentence length in the dataset and should be adjusted experimentally.

\subsection{Encoding Additional Information}
\label{sec:method-append}

Further modifications to the base models are related to the utilisation of information beyond sentence text. For the Goodreads dataset, multiple additional features are available: review author, book identifier, and book genres. The probability of spoiler varies greatly depending on the book genre. Thus, we proposed two approaches to leveraging the books genre information.

\subsubsection{Encoding Genres' Votes Distribution} \label{sec:method-genres_embedding}

There are 10 genre groups in the datasets (see \cref{tab:genre-spoilers}). Metadata for a single book contains the number of votes for each genre group. The most straightforward way to encode this information is a votes distribution, i.e. using the percentage of votes for genre group $i$ as value for item $i$ of the vector (see Figure~\ref{fig:genre_encoding}). The resulting vector is fed to the linear layer, concatenated with pre-trained model output, and then passed to final linear and dropout layers.

\begin{figure}[!htb]
    \centering
    \includegraphics[width=0.35\linewidth]{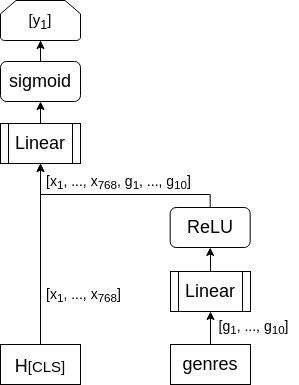}
    \caption{Schema of classification head utilizing concatenated genres' embeddings. $H_{[CLS]}$ represents the output of the BERT last encoder layer for \texttt{[CLS]} token; $Linear$ symbolises a linear layer; $sigmoid$ and $ReLU$ are activation functions; $y_1$ is the probability of the positive class.
    \label{fig:bert-genres_embedding}}
\end{figure}

\subsubsection{Appending Genres' Listing} \label{sec:method-genres_append}

The second approach leverages the textual aspect of genre information: genres for a particular book are appended to a review sentence (a model input) ordered by a descending number of votes. Figure~\ref{fig:genre_encoding} presents an example of the transformation. Some genres' group names were changed for easier-to-parse equivalents in order to avoid negations and long names, e.g. \textit{non-fiction} to \textit{fact} or \textit{history, historical fiction} to \textit{history}. This approach is far easier to implement for both single sentence and multiple sentences inputs as it does not require any architecture modification.

\begin{figure}[!htb]
    \centering
    \begin{subfigure}{\textwidth}
        \centering
       \textit{"sentence": "This is a great book!",\\ "genres": \{"comics, graphic":900,"young-adult":80,"fiction":20\}}
       \caption{The review text and the number of votes per genre group.}
    \end{subfigure}
    \begin{subfigure}{\textwidth}
        \centering
       \vspace{0.2cm}
        \textit{[0, 0.9, 0, 0.02, 0, 0, 0, 0, 0, 0.08]}
        \caption{Genre votes vector.}
    \end{subfigure}
    \begin{subfigure}{\textwidth}
        \centering
       \vspace{0.2cm}
        \textit{This is a great book! comics, youth, fiction}
        \caption{Textual input with encoded genres.}
    \end{subfigure}
    \caption{Encoding genre information in input data.}
    \label{fig:genre_encoding}
\end{figure}


\section{Experiments and Results}
\label{sec:experiments}

Our experiments were conducted for the spoiler datasets (from Table~\ref{tab:datasets-summary}), the dataset Goodreads-balanced was used only to fine-tune parameters, final evaluation was made on the full Goodreads dataset. At the end of this section we summarized all the tests and their results in Table~\ref{tab:tests-listing}.

\subsection{Training Performance}

To quantify the increase in training speed from leveraging TPU architecture, we compared the training time of the BERT-based model on different hardware architectures with the TV Tropes Movies dataset. Table~\ref{tab:hardware_speed_comparison} shows the total number of seconds required to complete a single training epoch on GPU (Tesla K80) and TPU (v2-8). 
For the same batch size, TPU finishes calculations more than 10 times faster than the GPU. Having the larger RAM available, TPUs further benefit from setting the batch size to the highest setting possible (512).

\begin{table}[htb]
\centering
\begin{tabular}{ccc}
\toprule
Batch size & {GPU epoch {[}s{]}} & {TPU epoch {[}s{]}} \\ \midrule
32         & 482               & 37                \\
512        & {-}               & 13                \\ \bottomrule
\end{tabular}
\caption{The time, in seconds, required for a single training epoch on TV Tropes Movies dataset for different hardware: GPU (Tesla K80) and TPU (v2-8).}
\label{tab:hardware_speed_comparison}
\end{table}

\subsection{TV Tropes Movies}
\label{sec:TVTropesMovies}
The TV Tropes Movies dataset contains only single sentence input, so we tested the SSC variant. We examined multiple Transformers models to serve as the basis for testing: BERT, ALBERT, and ELECTRA~\citep{clark_electra_2020}. Input sentences were truncated to 128-word pieces because 99.87\% of sentences are shorter than this limit. With a fixed batch size to 512, the best learning rate was 4e-05. The loss function used was binary cross-entropy weighted by the ratio of samples belonging to the positive and negative class in the train data: 
$\approx 0.9$ weight for the positive class. We evaluated the models on the original TV Tropes Movies dataset splits (see~\citet{boydgraber_spoiler_2013}): test set, we used development set \#1 as validation one and disregarded the development set \#2 to avoid confusion.

\begin{table}[!htb]
\centering
\begin{tabular}{rcccc}
\toprule
Model base                 & 
\begin{tabular}[c]{c}Learning \\ rate \end{tabular}   & 
\begin{tabular}[c]{c}Batch \\ size \end{tabular} & 
Accuracy & 
\begin{tabular}[c]{c}ROC \\ AUC \end{tabular} \\ \midrule
BERT Base-uncased          & 4e-05 & 512        & 0.7054   & 0.7985 \\
ALBERT Large-V2            & 2e-05 & 256        & 0.7355   & 0.8285 \\ 
ELECTRA \small{Base-discriminator} & 4e-05 & 512        & 0.7411   & 0.8042 \\ \bottomrule
\end{tabular}
\caption{Hyperparameters search results on TV Tropes Movies validation dataset for different pre-trained model serving as base.}
\label{tab:tvtropes_movies-val}
\end{table}

Although ALBERT and ELECTRA results for validation set were similar (Table \ref{tab:tvtropes_movies-val}), we selected the latter for the final testing also due to faster training time: 13 seconds vs. 30 for a single epoch.

\begin{table}[!htb]
\centering
\begin{tabular}{rcc}
\toprule
Model        & Accuracy & {ROC AUC}  \\ \midrule
SVM~\citep{boydgraber_spoiler_2013}          & 0.67     & {not reported}     \\
HAN-attempt                                 & 0.7034   & 0.7663   \\
HAN~\citep{wan_fine-grained_2019}            & 0.720    & 0.783   \\
SpoilerNet~\citep{wan_fine-grained_2019}     & 0.737    & 0.803   \\
Our ELECTRA-based model                     & 0.7406   & 0.8120  \\ \bottomrule
\end{tabular}
\caption{Results on the TV Tropes Movies test dataset. The values for metrics are reported as presented in cited papers.}
\label{tab:tvtropes_movies-test}
\end{table}

The final results (Table \ref{tab:tvtropes_movies-test}) show a slight advantage of our approach in comparison to previous methods. The GRU-based model presented in~\citet{chang_deep_2018} has a higher accuracy of 0.7556 but uses additional genre information as opposed to our text-only model. 

We attempted to recreate the modification of the HAN model described in~\citet{wan_fine-grained_2019} (\texttt{HAN-attempt}). However, the authors did not open their source code and reported a different division of the dataset to train/test/validation splits than in original dataset paper, which makes it impossible to reproduce the the exact result.

\subsection{TV Tropes Books}
The characteristics of the TV Tropes Books dataset are similar to the movies variant, so with the same assumptions, we made further tests to compare the two input variants: the SSC and the SeqSC approaches. 
Input sentences were truncated to 96-word pieces for the single sentence variant, because 99.76\% of sentences are shorter than this limit. For the SeqSC, we used five as the maximum number of sentences fed to the model at once, and we set the maximum input length to 512, the upper limit for BERT-like models. The input length determines the models' RAM usage, so the batch size had to be changed \\ accordingly. We also adjusted the learning rate according to the rule: decreasing batch size $\alpha$ times also decreases the learning rate $\alpha$ times~\citep{smith_dont_2018}. The dataset was split into three parts: training (80\%), validation (10\%), and testing (10\%) with stratification to guarantee the same distribution of classes on the document level in each of the splits.

\begin{table}[!htb]
\centering
\begin{tabular}{rcccc}
\toprule
Model                       & {Batch size} & {Learning rate} & {ROC AUC} & {PR AUC} \\ \midrule
SSC ELECTRA      & 512          & 4e-05           & 0.8463    & 0.5754   \\
SeqSC ELECTRA  & 128          & 1e-05           & 0.8429    & 0.5364   \\
\bottomrule
\end{tabular}
\caption{Results on the TV Tropes Books validation dataset.}
\label{tab:tvtropes_books-val}
\end{table}

Table~\ref{tab:tvtropes_books-val} indicates that the SeqSC model introduced no improvement over the SSC variant. It is worthy to notice, that the average number of sentences in a single document was 1.96. However, if we extended this dataset by longer reviews, the modeling results could occur better for the sequential model.

Table~\ref{tab:tvtropes_books-test} shows that our model achieved better results on the test data than the recreated HAN. 

\begin{table}[!htb]
\centering
\begin{tabular}{ccc}
\toprule
Model  & {ROC AUC} & {PR AUC} \\ \midrule

HAN-attempt        & 0.7932  & 0.4628 \\
SeqSC ELECTRA      & 0.8471  & 0.5706 \\ \bottomrule
\end{tabular}
\caption{Results on the TV Tropes Books test dataset.}
\label{tab:tvtropes_books-test}
\end{table}

\subsection{Goodreads}
In our tests, we used the balanced version of Goodreads data because of available computation limits. Both the full and balanced variants were split into three parts -- train (80\%), validation (10\%), and test (10\%) -- with stratification to guarantee the same distribution of classes on the document level in each of the splits. Assessment of the variants considered in each experiment is performed on the validation split of the balanced dataset.

\paragraph{Single-Sentence Model.} Input sentences were truncated to 96-word pieces because 99.91\% of the sentences were shorter than this limit. With the batch size fixed to 512, the best learning rate was 3e-05. The base loss function used was binary cross-entropy weighted by the ratio of samples belonging to the positive and negative class in the train data: 
$\approx 4.62$ weight for the positive class. 

The uncased base BERT model was a starting point for the experiments. We experimented with additional layers placed on top of the BERT model. We compared the following variants: using fully connected linear layer from pre-training and then taking the \texttt{CLS} token embedding from the last hidden layer (pooled model) vs. the omitting pre-trained layer (sequence model). However, the results were comparable with the ROC AUC 82.15\% for the sequence version and 82.16\% for the pooled one. Thus, we decided to discard the pre-trained linear layer due to its additional complexity without significant improvement in the metrics.

\paragraph{Pre-Trained Model Type.} Recent research proposes multiple variants of the Transformers model. We chose those with the same input-output characteristics to avoid different tokenization schemes. The BERT model is available in two variants: cased and uncased. The former preserves true case and accent markers of words, the latter changes all words to lowercase and strips out the accents. 
We tested both assuming the named entities may play a role in detecting spoilers. Additionally, we explored the ALBERT model because of its parameter efficiency and the ELECTRA model because of its relative novelty and innovative pre-training procedure. 

\begin{table}[!htb]
\centering
\begin{tabular}{rccc}
\toprule
Pre-trained model           & {Parameters {[}M{]}} & {ROC AUC} & {PR AUC} \\ \midrule
BERT-Base uncased          & 110                & 0.8215  & 0.5510 \\
BERT-Base cased            & 110                & 0.8206  & 0.5504 \\
ALBERT-Large-V2            & 17                 & 0.8242  & 0.5619 \\
ELECTRA-Base               & 110                & 0.8261  & 0.5606 \\ \bottomrule
\end{tabular}
\caption{Performance of different pre-trained models used as a base.}
\label{tab:exp-pretrained_model_type}
\end{table}

We selected the ELECTRA model for future experiments, aware of its highest results (ROC AUC 82.61\% and PR AUC of 56.06\%) and the longer training time of the ALBERT model. Then, we tested different loss functions. However, our results did not indicate any improvement for other than binary cross-entropy weighted by the ratio of minority class to the majority one. 

\paragraph{Loss Function.} The document-balanced version of the Goodreads dataset still has a disproportion between classes on sentence-level -- only 18\% of them contain spoilers. Using the standard binary cross-entropy as a loss function led to model always predicting majority class as expected. In this experiment, we aimed to test the alternatives: weighted binary cross-entropy and focal loss.

\begin{table}[!htb]
\centering
\begin{tabular}{rcc}
\toprule
Loss function & {ROC AUC} & {PR AUC} \\ \midrule
Focal loss    & 0.8260  & 0.5592 \\
Weighted BCE  & 0.8261  & 0.5606 \\ \bottomrule
\end{tabular}
\caption{Results for different loss function used to tackle class imbalance.}
\label{tab:exp-loss_function}
\end{table}

Results from~\cref{tab:exp-loss_function} does not indicate any improvement for the focal loss over binary cross-entropy weighted by the ratio of minority class to the majority one. As a consequence, we chose the latter for being more standard measure.

\paragraph{Book Genre Encoding.} \label{sec:genre_encoding-results}
Each review in the Goodreads dataset contains a book identifier, which can be used to add book genre information. As the spoiler percentage is different for every genre group, it may be highly beneficial to take into account book genre information. We tested two variants of including such metadata in the model: concatenating the sentence embedding from a pre-trained model with genres encoding and appending the genres' listing to the end of each sentence.

\begin{table}[!htb]
\centering
\begin{tabular}{rcc}
\toprule
Genre encoding        & {ROC AUC} & {PR AUC} \\ \midrule
None                   & 0.8261  & 0.5606 \\
Distribution embedding & 0.8268  & 0.5619 \\
Sentence text append   & 0.8311  & 0.5695 \\ \bottomrule
\end{tabular}
\caption{Results for different methods of incorporating book genres' metadata.}
\label{tab:exp-genres_encoding}
\end{table}

Both approaches leverage the basic variant, as seen in Table~\ref{tab:exp-genres_encoding}. The higher performance of the text append method can be attributed to the place of inserting metadata to the model: performed as early as possible allows the neural network to utilize all its parameters when processing this information. The distribution embedding approach may be too simplistic and applied too late to affect the embedding of the sentence significantly.

\paragraph{Sequential-Sentence Model.} The SeqSC model, introduced in~\citet{cohan_pretrained_2019}, utilizes the sentence context while performing a classification. We applied the sentences splitting method and tested a different number of sentences fed to the model at once. The learning rate used was changed to 1e-05 due to the lower batch size of 128.

\begin{table}[!htb]
\centering
\begin{tabular}{ccc}
\toprule
{Context size} & {ROC AUC} & {PR AUC} \\ \midrule
{None}         & 0.8311  & 0.5695 \\
3              & 0.8424  & 0.5665 \\
5              & 0.8437  & 0.5669 \\ 
7              & 0.8363  & 0.5531 \\ \bottomrule
\end{tabular}
\caption{Results for different maximum number of sentences (a context size) used as model input.}
\label{tab:exp-ssc_context_size}
\end{table}

Taking into account sentence neighbourhood engendered significant improvements, regardless of the additional context size 
(Table~\ref{tab:exp-ssc_context_size}). The results indicated the importance of closest sentences over more distant ones; the use of five sentences at once achieved the best performance. This variant was later combined with the genres' text append approach, by adding genres' listing at the end of combined sentences text, which resulted in another improvement to 84.52\% ROC AUC and 57.00\% PR AUC.

\paragraph{Final Evaluation.} The best performing combination so far consists of
\begin{inparaenum}
\item discarding the pre-trained additional linear layers,
\item using the ELECTRA model as the basis,
\item evaluating the prediction with weighted cross-entropy loss function,
\item appending the genre information to the sentence text, and
\item inputting five sentences at once to the model.
\end{inparaenum}
Previous experiments were performed on validation split of the balanced variant of the Goodreads dataset (due to constrained computational resources). Moreover, for a more reliable comparison with~\citet{wan_fine-grained_2019}, we conducted the final training and evaluation using the full Goodreads dataset. The only required adjustment was setting the weight parameter of binary cross-entropy loss to 
$\approx 29.96$: the ratio of samples belonging to the positive and negative class in the training data.

\begin{table}[!htb]
\centering
\begin{tabular}{rc}
\toprule
Model            & {ROC AUC} \\ \midrule
SVM              & 0.744   \\
CNN              & 0.777   \\
HAN              & 0.901   \\
SpoilerNet       & 0.919   \\
Our final model  & 0.8821   \\ \bottomrule
\end{tabular}
\caption{Results on test set of the Goodreads full dataset. The ROC AUC for other models is reported as in~\citet{wan_fine-grained_2019}.}
\label{tab:goodreads-final_evaluation}
\end{table}

Table~\ref{tab:goodreads-final_evaluation} shows that our approach performs slightly worse than other architectures proposed in~\citet{wan_fine-grained_2019}; however, we were unable to reproduce the results with HAN (see Section~\ref{sec:TVTropesMovies}). 

\subsection{Tests Summary}

A list of all tests is presented in Table~\ref{tab:tests-listing}.

The most straightforward way to improve the proposed model is to pre-train it on a task-related dataset in an unsupervised way before fine-tuning on a smaller dataset. This way, the model trained on a larger dataset can be utilized to aid the training on another one, when annotations are not present or not readily available. This approach was proved to be effective by~\citep{sun_how_2020}. 

\begin{table*}[tb]
\centering
\begin{tabular}{p{0.3cm}p{4.3cm}p{2.7cm}p{4cm}p{1cm}}
\toprule
No. & Test Scope  & Dataset & Selected Model/Parameters & AUC [\%] \\ 
\midrule
I. & Hyperparameters search results for different pre-trained model serving as base  & TV Tropes Movies develop set \#1 & ELECTRA Base discriminator (learning rate=4e-05, batch size=512); significantly faster training time & 80.42  \\
II. & Comparing with published results & TV Tropes Movies test set & ELECTRA as above 
& 81.20 \\
III. & Input variants: single and sequential sentences & TV Tropes Books valid set & ELECTRA single-sentence (as above) 
& 84.63\\
IV. & Comparing with a reproduced HAN model & TV Tropes Books test dataset & ELECTRA as above
& 84.71 \\
\hline
V. & Heads (pooled
, sequence
) placed on top of the BERT & document-balanced  Goodreads valid set & pooled but rejected -- not a significant improvement and more complexity & 82.16 vs. 82.15\\
VI. & Base of pre-trained model & as above & ELECTRA base & 82.61 \\
VII. & Loss function to tackle class imbalance & as above & weighted BCE  & 82.61 \\
VIII. & Methods of incorporating book genres metadata & as above & appending the genres listing to the input text & 83.11 \\
IX. & A number of sentences fed to the model at once (learning rate=1e-05,  batch size=128) & as above & 5 sentences at once & 84.37 \\
X. & Adding genres to IX. & as above & 5 sentences input and genres appending & 84.52 \\
\hline
XI. & Final test with full Goodreads training & test for  Goodreads full set & & 88.21 \\
\bottomrule
\end{tabular}
\caption{Scheduled tests and their best results, given in ROC AUC metric}
\label{tab:tests-listing}
\end{table*}

\section{Models Analysis}
\label{sec:analysis}

To understand the performance of the proposed models, we conducted a quantitative and qualitative analysis of their behaviour. We looked into their components and studied the reactions to the input perturbation. 

\paragraph{Genre Attention.} To analyse the improvements in results for the models with added book genre information (presented in~\cref{sec:genre_encoding-results})
, we focused on inspecting the attention mechanism. We only attempted to improve the transparency of the model, not interpret it (to avoid pitfalls of using the attention as a way of model interpretation). 

We studied the last two layers of the model because it is the part that undergoes the most significant change during the fine-tuning process, as observed in~\citet{kovaleva_revealing_2019}. Specifically, we looked at attention used for building the \texttt{[CLS]} token embedding with additional preprocessing steps, while contribution from itself and \texttt{[SEP]} tokens were omitted so as to emphasize word-level values; later, attention values were normalized.

\begin{table}[!htb]
\centering
\begin{tabular}{rrcc}
{Layer} & Class       & {Baseline} & {Real} \\ \midrule
12      & both        & 0.2709               & 0.2095 \\
        & non-spoiler & 0.2720               & 0.2114 \\
        & spoiler     & 0.2659               & 0.2000 \\ \cmidrule(lr){2-4}
11      & both        & 0.2709               & 0.1838 \\
        & non-spoiler & 0.2720               & 0.1871 \\
        & spoiler     & 0.2659               & 0.1679 \\ \bottomrule
\end{tabular}
\caption{
The part of normalized attention given to genre tokens when processing the \texttt{[CLS]} token. 
The values in reality (real) are compared to the baseline, which assumes a uniform distribution of attention across all tokens (uniform). Differences in baseline values originate from the various average tokens length in sentences within the specific class. The results are reported for 10\% of the data sampled randomly. 
}
\label{tab:attention-genre_tokens}
\end{table}

Table~\ref{tab:attention-genre_tokens} shows a part of attention dedicated to input genre data -- not text token themselves -- which is a mean for all input samples from all the tested examples. The baseline is counted as if the model paid attention to all tokens with the same weight: the text itself and genre tokens as well. The values of mean attention for the real model are lower than in baseline settings. Thus, we may conclude that the tested model (ELECTRA) paid more attention to sentence text and token themselves, while less to the genre.  Nevertheless, the genre is also considered but with less weight.  We interpret it as the proper working of the model. 

\paragraph{Spoiler Words Masking.} The \textit{TV Tropes Books} dataset contains word-level spoiler annotations. We used them to analyze model behaviour. We selected 34,000 sentences with partial spoiler tags, which covered less than 90\% of sentences. The sentences with a single spoiler tag constitute the majority of all sentences (89\% of all partial spoilers) and are easier to compare. The following experiments were inspired by~\citet{deyoung_eraser_2020}.

\begin{figure}[!htb]
    \centering
    \includegraphics[width=.5\linewidth]{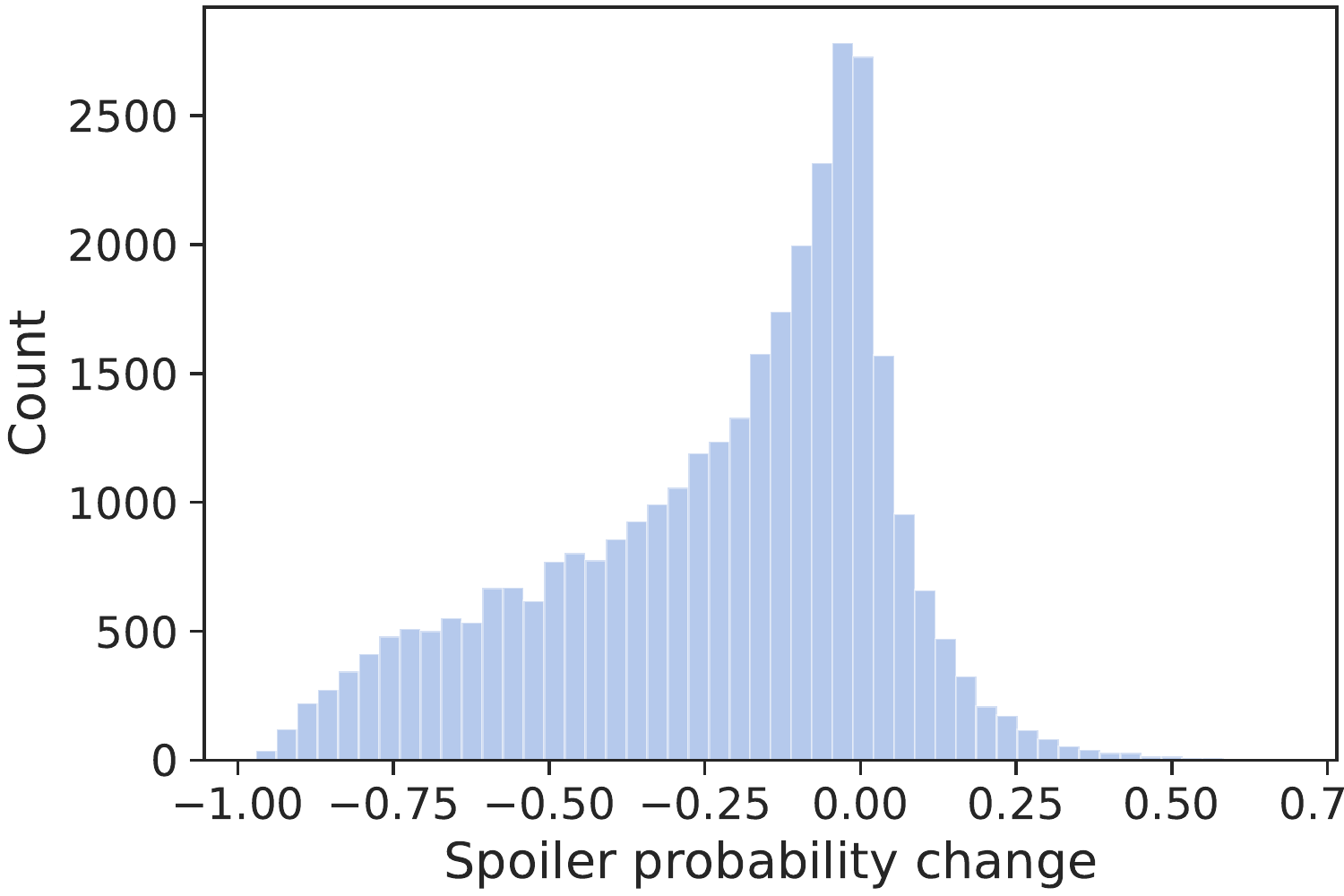}
    \caption{Comprehensiveness test results: changes in the estimated probabilities (model outputs) after removing the spoiler part of the input sentences.}
    \label{fig:remove_spoiler-spoiler_probability_change}
\end{figure}

\paragraph{Comprehensiveness.}
In the comprehensiveness experiment, we observed the difference between probabilities predicted for the original sentence and the sentence with the spoiler words removed. If the spoiler annotation is \textit{comprehensive}, the model should be less confident about the prediction after the removal of the spoiler part. In other words, assuming the annotations and model are perfect, there should be a considerable drop in spoiler probability for every single input data. Figure~\ref{fig:remove_spoiler-spoiler_probability_change} summarizes the probability changes after the removal of spoiler parts. In 81\% of cases, removing the spoiler part of the input sentence resulted in a lower probability of spoiler class estimated by the model.

The main difference between cases with increased and decreased spoiler probability was the location of impactful verbs.  When such a verb is inside a spoiler tag -- and it is removed -- we noticed lower spoiler probability; if it remains outside the spoiler tag, the probability raises or remains unchanged (see also~\citep{boydgraber_spoiler_2013}). They are connected with Transitivity concept from~\citep{hopper_transitivity_1980,boydgraber_spoiler_2013} in the context of the spoiler detection task. 

\begin{figure}[!htb]
    \centering
    \includegraphics[width=.5\linewidth]{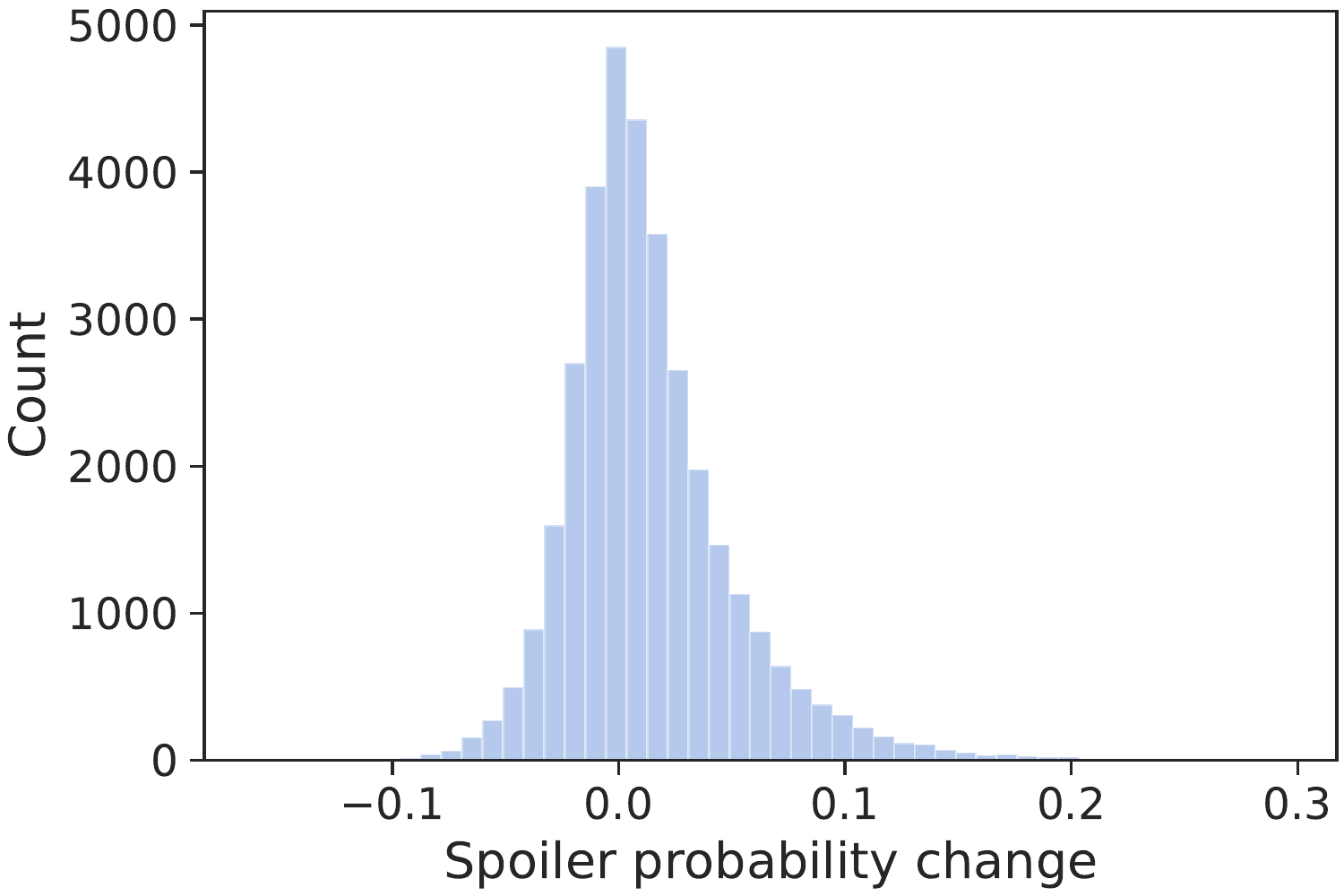}
    \caption{Sufficiency test results: changes in the estimated probabilities (model outputs) after leaving only the spoiler part of the input sentences.}
    \label{fig:leave_only_spoiler-spoiler_probability_change}
\end{figure}

\paragraph{Sufficiency.}
In the next experiment, we evaluated the model's \textit{sufficiency}. The modified sentence consisted only of the spoiler part, and we compared its spoiler probability with the original one (change in model outputs). The output for the modified sentence was higher in 61\% of the examples (Figure~\ref{fig:leave_only_spoiler-spoiler_probability_change}). In 78\% of cases, probability did not drop below 90\% of the original estimation.

The highest changes were often observed for the input cases when the spoiler contained a named entity. However, the positive changes often did not contain any additional words -- they consisted only of the named entity. Table~\ref{tab:leave_only_spoiler-named_entities} quantifies this observation.

\begin{table}[htb]
\centering
\caption{Percentage of spoilers containing a named entity and consisting only of the named entity. "Probability change" column refers to a sign of the spoiler probability change after leaving only the spoiler part of the sentence; $N$ is a number of samples with most significant probability taken into consideration.}
\label{tab:leave_only_spoiler-named_entities}
\begin{tabular}{p{2cm}p{1.5cm}p{3.5cm}p{3.5cm}}
\toprule
Probability change   & N    & {Contains named entity {[}\%{]}} & {Is a named entity {[}\%{]}} \\ 
\midrule
Positive & 1000 & 65.30                  & 46.30            \\
Negative & 1000 & 62.90                  & 15.20            \\
Positive & 100  & 69.00                  & 53.00            \\
Negative & 100  & 65.00                  & 28.00            \\ 
\bottomrule
\end{tabular}
\end{table}

\section{Conclusions}

Our new dataset contains word-level spoiler annotations that can be used in multiple ways:
\begin{inparaenum}[(i)]
  \item to perform word-level classification,
  \item to analyse and interpret the models with common ground-truth,
  \item to learn how to extract or generate explanations for model predictions,
  \item to provide a base for defining new tasks related to spoilers beyond classification, e.g. to extract plot information from books comments.
\end{inparaenum}

The methods and models we used are not novel. We researched how to adapt them to the spoiler task, which is a binary classification problem.
\footnote{Our code for building models and their analysis: \url{https://github.com/rzepinskip/spoiler-detection}, reproducing the HAN model for spoiler datasets: \url{https://github.com/rzepinskip/han-spoiler}.} Moreover, attention analysis and interpretability techniques used come from previous research on different text sets and tasks. However, we adopted the research and proved that we could gain more insights on new models and datasets by using those techniques.

The most straightforward way to improve the proposed model is to pre-train it on a task-related dataset in an unsupervised way before fine-tuning on a smaller dataset. This way, the model trained on a larger dataset can be utilized to aid the training on another one, when annotations are not present or not readily available. This approach was proved to be effective by~\citet{sun_how_2020}. 

The idea for and the systematic method of providing explanations alongside predictions are exciting approaches to gain users' and stakeholders' trust when developing an automated text detection system. That feature is highly desirable given the complicated nature of spoilers.


\bibliographystyle{unsrtnat}
\bibliography{references}  

\end{document}